\documentclass{article}
\usepackage{spconf,amsmath,epsfig,graphicx,cite}
\usepackage{subfigure}
\usepackage{subfig}
\usepackage{color}
\usepackage{graphicx} %use graph format
\usepackage{epstopdf}

\begin{document}\sloppy
% \bibliographystyle{IEEEbib}
% Example definitions.
% --------------------
\def\x{{\mathbf x}}
\def\L{{\cal L}}

% \name{Anonymous ICME submission}
% \address{}
% Single address.
% --------------
% Title.
% ------
\title{Correlation filter tracking with adaptive proposal selection for accurate scale estimation}

\name{Luo Xiong, Yanjie Liang, Yan Yan, Hanzi Wang}
\address{
Fujian Key Laboratory of Sensing and Computing for Smart City,\\
School of Information Science and Engineering, Xiamen University, Xiamen, China\\
xiongluo@stu.xmu.edu.cn, yanjieliang@yeah.net, \{yanyan, hanzi.wang\}@xmu.edu.cn
}
\maketitle
% \address{}
% \{xiongluo,qiangwu\}@stu.xmu.edu.cn, yanjieliang@yeah.net,\\ @stu.xmu.edu.cn, \{yanyan, Hanzi.Wang\}@xmu.edu.cn}

% Many trackers cope with scale variation with fixed scale instead of generating aspect ratio scale proposals from the target appearance.
%
\begin{abstract}
Recently, some correlation filter based trackers with detection proposals have achieved state-of-the-art tracking results. However, a large number of redundant proposals given by the proposal generator may degrade the performance and speed of these trackers. In this paper, we propose an adaptive proposal selection algorithm which can generate a small number of high-quality proposals to handle the problem of scale variations for visual object tracking. Specifically, we firstly utilize the color histograms in the HSV color space to represent the instances (i.e., the initial target in the first frame and the predicted target in the previous frame) and proposals. Then, an adaptive strategy based on the color similarity is formulated to select high-quality proposals. We further integrate the proposed adaptive proposal selection algorithm with coarse-to-fine deep features to validate the generalization and efficiency of the proposed tracker. Experiments on two benchmark datasets demonstrate that the proposed algorithm performs favorably against several state-of-the-art trackers.
\end{abstract}
\section{Introduction}
\label{sec:intro}

% Real-time visual object tracking plays an important role in many applications in computer vision, such as video surveillance, human-computer-interfaces, robotic analysis and automation, etc. Generic object tracking is a challenging task of predicting a trajectory of a generic object in a video sequence, only initialed by the first frame.
% There are many challenges in this field, such as scale variation, motion blur, illumination change and occlusion, etc.
Visual object tracking plays an important role in many applications of computer vision, such as video surveillance, human-computer interface and robotic analysis. One of the main challenges of object tracking is to handle the scale variations of targets caused by deformation, fast motion and rotation, etc.

% MOSSE firstly proposes the correlation filter in the frequency domain into this field. And KCF proposed by Henriques et ral. utilizes the Gaussian kernel into the convolution operation~\cite{henriques2015high}.

In recent years, the correlation filter based trackers~\cite{henriques2015high,han2018content} have attracted much attention due to their high efficiency and accuracy. These trackers mainly learn the correlation filters in the Fourier domain to detect the most likely candidate in each frame. In particular, the convolution operation with the correlation filters in the spatial domain corresponds to the element-wise operation in the Fourier domain, leading to the high efficiency of object tracking.

% And the highest peak from convolution output in spatial domain usually indicates the location and the confidence between the correlation filter model and the candidate.The circulant matrix can be roughly considered as the dense sampling.
% Though many state-of-the-art trackers can cope with those sequences very efficiently, the bounding boxes of fixed sizes are not able to handle large scale change and appearance deformation caused by fast motion[3BACF].
There are two main ways to address the scale variations of targets during tracking. One common way is to design several scales empirically or employ the extra correlation filters to select the best scale for each frame. For example, the HCF tracker~\cite{ma2015hierarchical} utilizes three layers from the deep network as three separate models. Then, it extracts the HOG features from the patches with different scales to decide the best scale for each frame. The SAMF tracker~\cite{li2014scale} copes with the scale variations by designing some scales and employing the correlation filters with different scales to obtain the final scale. DSST~\cite{danelljan2014accurate} and fDSST~\cite{danelljan2017discriminative} formulate a correlation filter based on the DCF tracker~\cite{henriques2015high} and another one-dimensional correlation filter to estimate the location and scale, respectively. These algorithms are based on the fixed aspect ratio of the ground-truth in the initial frame, and scales are empirically designed, thus resulting in sub-optimal tracking accuracy.

% They combine the proposals from EdgeBoxes with the correlation filter to locate and handle scale variation.

The second way is to combine the detection proposal generators with correlation filters to handle scale variations~\cite{liang2018robust,huang122015enable}. For instance, KCFDPT~\cite{huang2017applying} utilizes KCF~\cite{henriques2015high} to detect the initial location in each frame, where the EdgeBoxes algorithm~\cite{zitnick2014edge} is adopted to generate proposals. Owing to the effective combination of detection proposals and correlation filters, KCFDPT achieves more accurate tracking performance than fDSST and SAMF. However, the redundant proposals generated by EdgeBoxes hinder the efficiency of KCFDPT. Moreover, the distractors in the redundant proposals generated by the edge information may degrade the tracking performance in consecutive frames.

To solve the above problems, we propose an algorithm to adaptively select a small number of high-quality proposals and effectively handle the scale variations for correlation filter based trackers. Our contributions are summarized as follows:

\begin{itemize}
  \item We employ the color information to measure the similarity between the instances and proposals generated by EdgeBoxes. Based on the HSV color histogram, we propose an adaptive selection strategy to discard the redundant proposals based on the confidence of the current correlation filter. Therefore, we decrease the distractors in the redundant proposals and improve the tracker with better performance and faster speed.
  \item We further integrate the proposal selection algorithm with coarse-to-fine deep features derived from the VGGNet~\cite{simonyan2014very} to demonstrate the high quality of the selected proposals. Extensive experimental results on two benchmarks (OTB2013~\cite{wu2013online}, OTB2015~\cite{wu2015object}) demonstrate that the adaptive proposal selection algorithm effectively improves the tracking performance, especially in terms of scale variations.
\end{itemize}

\section{Baseline Algorithm}

In this section, we present a brief introduction of the KCFDPT~\cite{huang2017applying} algorithm, which is the baseline tracker in this paper. KCFDPT contains two parts: the KCF tracker~\cite{henriques2015high} and the proposal generator that uses EdgeBoxes~\cite{zitnick2014edge} with background suppression.

% utilizes EdgeBoxes equipped background suppression to generate proposals at the initial position. Finally, it employs KCF again in each proposal to select the proposal with the highest response.
% \noindent\textbf{Training stage by KCF}. 

In KCF, the objective of the correlation filter formulation is to learn a correlation filter $\omega$ and minimize the squared error over a set of samples $\{x_{1}, x_{2},..., x_{j},..., x_{n}\}$ and the corresponding regression targets $\{y_{1}, y_{2},..., y_{j},..., y_{n}\}$, where $j$ ranges from $1$ to $n$. $x_{j}$ is the $j$-th cycilc shift of the base sample $x_{1}$. $y_{j}$ is the corresponding label generated by a Gaussian function. $y_{1}$ is the label for the base sample $x_{1}$, which equals to 1. The problem can be written as:

% KCF computes the prediction function $f(z)=\omega^{T}z$, where $\omega$ is the parameter matrix of the correlation filter model and $z$ denotes an image patch used to train the model.

\begin{equation}
    \mathop{min}\limits_{w}\sum_{j}^{n}(f(x_{j})-y_{j})^{2} + \lambda\|\omega\|^{2},
\end{equation}
where $\lambda$ denotes a regularization parameter to alleviate the over-fitting problem. $n$ denotes the number of the training samples. Owing to the properties of the circulant matrix, the problem of Eq.~(1) has the closed-form solution, which is:
\begin{equation}
    \hat{\alpha} = \frac{\hat{y}}{\hat{k}^{x_{1}x_{1}} + \lambda},
\end{equation} where $\alpha$ represents the parameter matrix of the correlation filter in dual space, as opposed to $\omega$ in the primal space. The hat means the discrete Fourier transform. $k^{x_{1}x_{1}}$ represents the kernel correlation operation of the base sample $x_{1}$.

% \noindent\textbf{Detection stage in KCFDPT}.

Given an image patch $z$, KCF will be utilized to obtain the response map and detect the location of the target. The response map in KCF is computed as follows:
\begin{equation}
    \hat{f}(z) = \hat{k}^{\bar{x}z}\odot{\hat{\alpha}},
\end{equation} where $\hat{f}(z)$ is the response map of the image patch $z$ in the Fourier domain, whose maximum in the spatial domain indicates the detection location and confidence. $\hat{k}^{\bar{x}z}$ denotes the kernel correlation operation between the current appearance model $\bar{x}$ and the image patch $z$ in the Fourier domain. $\odot$ denotes the element-wise product.

KCFDPT utilizes EdgeBoxes~\cite{zitnick2014edge} to assign the background suppression weights to edges intersecting the boundary of the image patch and calculate the edge response value $r_{i}$ of each pixel $i$.
After obtaining the response value of each pixel from the background suppression factors, the score for a bounding box $b$ is evaluated by:
\begin{equation}
    h_{b} = \frac{\sum_{i\in{b}}c_{i}r_{i}}{2(b_{u}+b_{v})^{\kappa}} - \frac{\sum_{l\in{b^{in}}}r_{l}}{2(b_{u}+b_{v})^{\kappa}},
\end{equation}
where $r_{i}$ denotes the edge response value of a pixel $i$ within a bounding box $b$. $b_{u}$ and $b_{v}$ are the width and height of $b$, respectively. $b^{in}$ stands for the central region of $b$, whose size is $b_{u}/2 \times b_{v}/2$. $c_{i} \in{[0,1]}$ is a parameter, measuring how likely the contour that $i$ belongs to, is wholly contained in $b$. $\kappa$ is also a parameter to penalize the boxes with the large size.

In the detection stage, KCFDPT employs KCF to localize the center of the detection patch. Then, KCFDPT utilizes the parameters (e.g., intersection over union (IoU)) to select proposals generated by EdgeBoxes with the background suppression. The response map of each proposal can be obtained by Eq. (3), and the proposal with the highest response will be selected as the most promising proposal. Finally, the most promising proposal will be used to localize the target and update the correlation filter model $\alpha$ in Eq. (2).

Note that the proposals in KCFDPT are generated by using the EdgeBoxes algorithm, which only considers the edge information in the detection patch. Therefore, the proposals are not robust enough to handle motion blur and scale variations. Furthermore, the proposals contain the redundant information that may degrade the tracking performance.  In this paper, we propose to exploit the color information to adaptively select a small number of high-quality proposals to improve and accelerate the baseline  KCFDPT tracker.
% Firstly, KCFDPT utilizes the affinity of edges detected by EdgeBoxes~\cite{zitnick2014edge} to assign weights to pixels. Within the detection window $z_{d}$, the background suppression weight $\omega_{z_{d}}$ of edge groups is defined as:
%   \begin{equation}
%   \begin{aligned}
%     \omega_{z_{d}}&=\frac{1}{4}\mathop{max}\limits_{E}\prod\limits_{j=1}\limits^{|E|-1}a(e_{j}, e_{j+1})\\
%     a(e_{i},e_{j})&=|cos(\theta_{i} - \theta_{ij})cos(\theta_{ij} - \theta_{i})|^{2}
%      \end{aligned}
% \end{equation} Where $E$ denotes an ordered edge group path of the length $|E|$ starting from the edge group $t_{1} = s$ to the final edge group $t_{|T|}$ which intersects the boundary of the patch $z_{d}$. And $s$ represents the whole edge group. Function $a$ calculates the edge group affinity, which represents the similarity between the edge group $e_{i}$ and $e_{j}$. Here $\theta_{i}$ and $\theta_{j}$ are orientation angles of $e_{i}$ and $e_{j}$ on average, respectively. Then the edge response value $r_{i}$ of each pixel $i$ is calculated as:
% \begin{equation}
%     r_{i} = r_{i} \times{(1 - \omega_{z_{d}}(s))} , i \in s
% \end{equation}
% $\mathop{min}\limits_{w}$

\section{Our Algorithm}

The pipeline of the correlation filter based tracker with the adaptive proposal selection  is shown in Fig.~1.
During the tracking process, we firstly utilize the correlation filter based tracker to obtain the initial location and the target size for the current frame, as shown in Fig. 1(b). Then, EdgeBoxes with background suppression is adopted to generate detection proposals, which are further selected by the IoU constraint, as shown in Fig. 1(c). Next, based on the HSV color histograms (Section 3.1) of both the proposals and instances, the adaptive proposal selection (Section 3.2) is utilized to discard the redundant proposals, where the correlation filter model will select the proposal with the highest response, as shown in Fig. 1(d). Finally, the prediction for the current frame is a trade-off between the most promising proposal and the initial prediction given by KCF, as shown in Fig. 1(e). We further integrate the proposal selection algorithm into the deep features based correlation filters (Section 3.3) for robust object tracking.

% Finally, the current prediction for each frame is a tradeoff between the most promising proposal and the initial prediction given by KCF.
% The overview of the proposed proposal selection algorithm is given in Section 3.1 and 3.2. KCFDPT improved by our proposed algorithm is called CFAPS. In order to verify the generalization and efficiency of the proposal selection algorithm, we in tegerate the deep features with the whole proposal generation. The deep version with the proposal selection algorithm is called DeepCFAPS.

\begin{figure*}[ht]
    \centering
    \includegraphics[height=4cm,width=16cm]{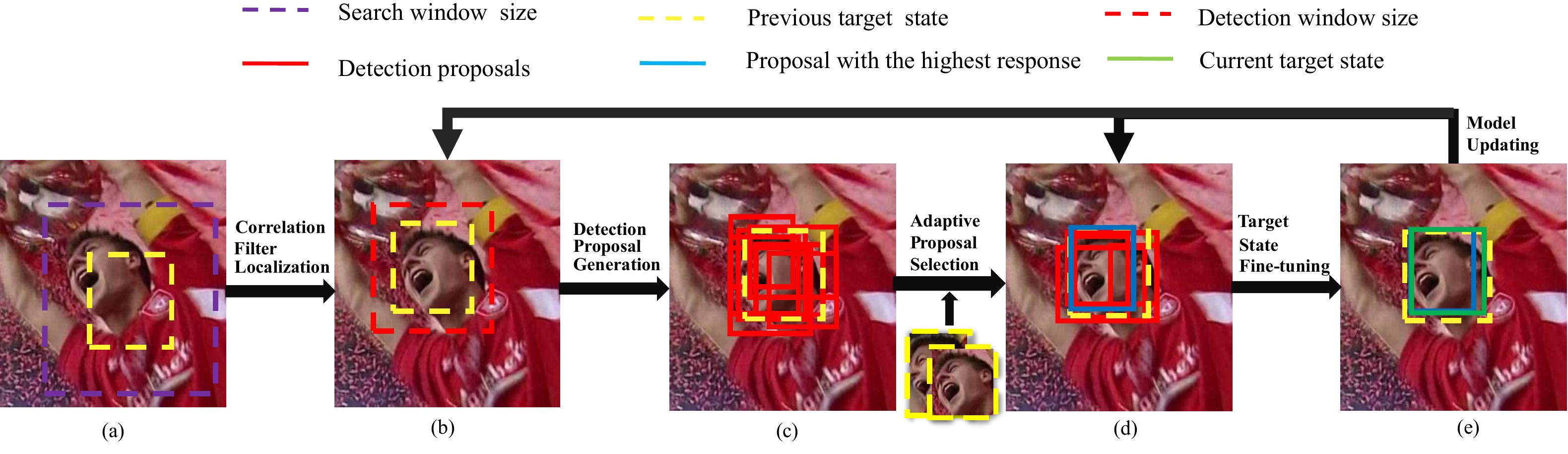}
    \caption{The pipeline of the correlation filter based tracker with the proposed adaptive proposal selection algorithm.}
    \label{fig:my_label}
    \vspace{-3mm}
\end{figure*}

\subsection{Color similarity measurement}

When the $i$-th frame comes, the location $\mathbf{o}_{i-1}$ and target size ($w_{i-1},h_{i-1}$) from the previous frame will be used as the inputs of KCF to detect the initial predicted location $\mathbf{o}_{i}'$. We keep the instance $I_{i-1}$ from the previous frame with the location $\mathbf{o}_{i-1}'$ and size ($w_{i-1},h_{i-1}$).
% The predicted target from the previous frame with the location $\mathbf{o}_{i-1}$ and the size ($w_{i-1},h_{i-1}$) is regarded as the instance $I_{i-1}$. 
In this paper, the initial instance $I_{1}$ (i.e., the initial target in the first frame) with size ($w_{1},h_{1}$) in the first frame and the previous instance $I_{i-1}$ (i.e., the predicted target in the previous frame) are two instances that we maintain during the tracking process. The EdgeBoxes with background suppression algorithm is performed on a detection window patch $z_{d}$. The center location and size of $z_{d}$ are $\mathbf{o}_{i}'$ and ($s_{d}w_{i-1}, s_{d}h_{i-1}$). $s_{d}$ is a parameter to render the detection window slightly larger than the previous size ($w_{i-1},h_{i-1}$).
%The detection window is indicated by the 'Detection size' in Fig.1b.
The output of EdgeBoxes with background suppression contains the redundant proposals, as visualized in Fig.~1(c). Based on these proposals, we further select the proposals which are similar to these two instances, namely, $I_{i-1}$ in the previous frame and $I_{1}$ in the first frame.

% After obtaining the background suppression proposals, we employ the color information to select the high-quality proposals.

% Therefore, we propose the proposal selection algorithm to obtain a small number of high-quality proposals

% a detection window patch $z_{d}$ will be extracted, whose center is the initial position $pos_{i}^{I}$ and size is $s_{d}sz_{i-1}$ in the current frame. EdgeBoxes and the background suppression operation mentioned in Section 2.2. are utilized to compute the edge response of each pixel in $z_{d}$. And the IoU between each proposal and instance $I_{i}$ is used to select proposals.

% After acquiring the background suppression score for each pixel, the whole patch is traversed in a sliding window manner and the scores of aspect ratio bounding boxes will be calculated. The bounding boxes whose scores are higher than $minScore$ will be selected and filtered by non-maximal suppression(NMS). And then these boxes whose IoU with the current instance are in a range $(r_{1},r_{2})$ will be saved.

% After obtaining the background suppression proposals, the aspect ratio proposals are generated by edge information. We further employ color information to refine the proposals.

% \subsubsection{Generating histogram using HSV color space}

%We calculate the distances between two instances and the proposals generated by EdgeBoxes equipped with background suppression.Inspired by~\cite{smith1978color}, 

HSV color space is shown to have better results for image retrieval than RGB color space~\cite{sural2002segmentation}. The proposals and two instances are represented in the HSV color space. More specifically, for each pixel, we firstly normalize the three channels of HSV (i.e., hue, saturation, and value) into the range of $[0,255]$. Secondly, we synthesize the image of three channels into the image of one channel, whose values range from $0$ to $255$. Therefore, we uniformly quantify the values of the three channels into $16,4,4$ levels, respectively. Then, we multiply the three values of each pixel by $16,4,1$ and add them together. As a result, each pixel can be represented by 8 bits, where the first 4, middle 2 and last 2 bits stand for the $16,4,4$ levels of hue, saturation, and value, respectively.

Mathematically, the color histogram of each proposal or instance in the HSV color space can be obtained by:
\begin{equation}
\begin{aligned}
    H=Hist(16\times&{Q_{16}(P_{hsv}(m,n,1))}+\\
    4\times&{Q_{4}(P_{hsv}(m,n,2))}+\\
    &{Q_{4}(P_{hsv}(m,n,3))}),
\end{aligned}
\end{equation}
where $Q_{16}(P_{hsv}(m,n,1))$ represents that the pixel value in the $m$-th row and the $n$-th column in the first channel of the proposal $P$ is quantified into 16 levels. $Q_{4}(P_{hsv}(m,n,2))$ represents that the pixel value in the second channel of the proposal $P$ is quantified into 4 levels.
 $Q_{4}(P_{hsv}(m,n,3))$ represents the similar meaning for the third channel. $Hist$ stands for the counting procedure of generating the color histogram, which ranges from 0 to 255. The histogram of instance $I$ can be calculated similar to the proposal $P$. 

% \subsubsection{Measuring similarity by Bhattacharyya distance}

After obtaining the color histograms of two instances and the proposals in the current frame, we further employ the Bhattacharyya coefficient to measure the similarity in the HSV color space. The similarity between the instance $I$ ($I_1$ or $I_{i-1}$) and the proposal $P$ can be computed as follows,
\begin{equation}
Sim_{P}^{I} = \sum\limits_{r\in{R}}{\sqrt{\frac{H_{I}(r)}{N(I)}\frac{H_{P}(r)}{N(P)}}},
\end{equation}
where $R$ is the range of bins in the color histogram, which is $[0,255]$. $H_{I}(r), H_{P}(r)$ represent the color histogram vectors of the instance $I$ and the proposal $P$, respectively. $N(I), N(P)$ denote the number of pixels in the instance $I$  and the proposal $P$, respectively. In this way, the fraction denotes the normalization operation of color histogram. 
% \begin{equation}
% Sim_{P}^{I} = \sum\limits_{r\in{R}}{\sqrt{\frac{H_{I}(r)}{numel(I)}\frac{H_{P}(r)}{numel(P)}}},
% \end{equation}

\subsection{Adaptive proposal selection}
The instance $I_{i-1}$ may be contaminated during the tracking process, so we propose an adaptive proposal selection strategy to choose the informative proposals.
%In the correlation filter based trackers, a learning rate is usually set to learn the current model matrices and the previous model matrices. In this paper, w
For each frame, we update the mean confidence of correlation filters as, % which is the maximum value of the selected response map.obtain the instance $I_{i}$
\begin{equation}
f_{mean}^{i} = (1 - \eta) f_{mean}^{i-1} + \eta f_{max}^{i},
\end{equation}
where $f_{mean}^{i-1}$ is the mean confidence of correlation models from the previous frames. $f_{mean}^{i}$ is the corresponding confidence from the first frame to the current frame. $f_{max}^{i}$ denotes the maximum value of the response map of the final selected proposal in the $i$-th frame. $\eta$ is a model confidence factor.

For the $i$-th frame, KCF is employed to obtain the response map and localize the center of detection window. We compare the temporary maximum response value $f_{max}^{i'}$ from KCF with $f_{mean}^{i-1}$ to indicate whether the instance $I_{i-1}$ is reliable or not. When $f_{max}^{i'} < \eta^{'}f_{mean}^{i-1}$, it indicates the instance $I_{i-1}$ is likely to be contaminated and unreliable. $\eta'$ is a rate to find those instances with too small response values. Therefore, we count the number of contaminated frames $\Delta_{i}$ as,
\begin{equation}
    \Delta_{i} = \sum\limits_{t=i_{1}}^{i}{\frac{1}{2}(sign( \eta'f_{mean}^{t-1} - f_{max}^{t'})+1)},
\end{equation}
where $sign$ is a sign function and $i_{1}$ denotes the frame number of the previous confident instance.

The final score of the proposal $P$ can be calculated by:
\begin{equation}
S_{P} = (1-e^{(-\alpha_{D}\Delta_{i})})Sim_{P}^{I_{1}} +  e^{(-\alpha_{D}\Delta_{i})} Sim_{P}^{I_{i-1}},
\end{equation}
where $\Delta_{i}$ is the number of the contaminated frames and $\alpha_{D}$ is a trade-off parameter. $Sim_{P}^{I_{1}}$ and $Sim_{P}^{I_{i-1}}$ denote the color scores between the proposal $P$ and the instances $I_{1}$ and $I_{i-1}$, respectively. When $\Delta_{i}$ is larger, it indicates that the previous instance $I_{i-1}$ is not reliable enough to select proposals (the instance $I_{i-1}$ is contaminated with the high probability) and we should mainly rely on the uncontaminated instance $I_{1}$.

Based on Eq.~(9), we rank the proposals in the descending order according to the similarity between each proposal and two instances. Then we discard about half of proposals to remove the distractors and accelerate the tracker. The results of proposal selection can be visualized in Fig.~1(d). After we obtain these selected proposals, Eq.~(3) is used to obtain the response map of each high-quality proposal. The proposal with the highest response is the candidate proposal.
From Fig. ~1, we can see that the proposal selection algorithm is effective for the correlation filter based trackers.

To avoid the over-sensitive problem and reduce the estimation error, a damping factor $\beta$ is used to obtain the target state in the $i$-th frame and keep a balance between the initial prediction and proposal selection. Assume that the proposal $P$ with the location $\mathbf{o}_{i}^{P}$ and the size ($w_{i}^{P}, h_{i}^{P}$) is the final selected proposal. The target state fine-tuning process is formulated as follows,
\begin{equation}
\begin{aligned}
   \mathbf{o}_{i}  &= \mathbf{o}_{i}' + \beta(\mathbf{o}_{i}^{P}-\mathbf{o}_{i}'),\\
    (w_{i}, h_{i}) &= (w_{i-1}, h_{i-1}) + \beta((w_{i}^{P}, h_{i}^{P}) - (w_{i-1}, h_{i-1})),
\end{aligned}
\end{equation}
where $\mathbf{o}_{i}'$ and $(w_{i-1}, h_{i-1})$ denote the initial location by KCF and  the previous size . $\mathbf{o}_{i}$ and ($w_{i}, h_{i}$) are the final prediction in the $i$-th frame. The prediction will be used to localize the target and update the correlation filter model $\alpha$ in Eq. (2).

%implement experiments with the hand-crafted feature based version CFAPS and deep CNNs based version DeepKCFDPT\_HSV to validate the generalization and efficiency of the proposal selection algorithm. DeepKCFDPT\_HSV integrates KCF with deep coarse-to-fine features, which can be regarded using HCF without the scale scheme as the baseline tracker.

% In this section, we further integrate the HSV proposal selection algorithm with the deep coarse-to-fine features and verify the good quality of selected proposals. HCF will replace the correlation based tracker KCF in the whole pipeline.

% \subsubsection{Using coarse-to-fine features to find the initial location}As for different layers in deep features, we do not concatenate them directly like hand-crafted features.

\subsection{Integrating with deep features}
The baseline KCFDPT tracker~\cite{huang2017applying} integrates KCF with color naming, image intensity and HOG features. Since the proposed adaptive proposal selection algorithm is generic and can be combined with different correlation filter based trackers, we also integrate the coarse-to-fine deep features (as used in the HCF tracker~\cite{ma2015hierarchical}) with the proposed algorithm to verify the generalization of the proposed algorithm. 

 More specifically, when the $i$-th frame comes, the location $\mathbf{o}_{i-1}$ and target size ($w_{i-1}, h_{i-1}$) from the previous frame are used to extract different layers of features (i.e., coarse-to-fine features) from VGG-Net. Deep features contain more semantic information than shallow features, but the resolutions of the deep features and shallow features are different. Hence, the bilinear interpolation operation is utilized to ensure that the shallow and deep features can be fused together. Then, the response map of deep features can be obtained by,
\begin{equation}
    g(z) = \sum\limits_{d=1}\limits^{D}\mu_{d}\mathcal{F}^{-1}(\hat{\omega}_{d}\odot{\hat{z}_{d}^{*}}),
\end{equation} where $D$ is the number of layers of deep features we use. $\mu_{d}$ denotes the weight of the $d$-th layer and $\mathcal{F}^{-1}$ denotes the inverse discrete Fourier function. $\odot$ denotes the element-wise product. $\hat{\omega}_{d}$ stands for the $d$-th correlation filter model matrix in the Fourier domain. $\hat{z}_{d}^{*}$ represents the current feature of the patch $z$ in the $d$-th layer in the Fourier domain. And $*$ denotes the complex-conjugate operation. The maximum of $g(z)$ indicates the initial location in the current frame.

%  $\mathbf{o}_{i}'$ and the confidence $g(z)_{max}^{I}$.

After obtaining the initial location by Eq. (11), the EdgeBoxes algorithm with background suppression will be employed to generate some proposals (Section 2). Based on these proposals, half of them will be discarded according to the similarity between instances and proposals (Section 3.1.1 and 3.1.2). Finally, the proposals will be evaluated by~Eq. (11) to find the proposal with the highest response, and the final prediction in the current frame is determined by Eq.~(10) from the candidate  proposal and the initial prediction.

\section{Experiments}
 We perform comprehensive experiments on two benchmarks: OTB2013~\cite{wu2013online} and OTB2015~\cite{wu2015object}. And we also evaluate the tracking performance of different trackers under the  scale variation attribute. 
 
%  In this paper, we set up the comparison on the two benchmarks with the challenging attribute, scale variation.
% \subsubsection{Proposal selection by coarse-to-fine features}
\subsection{Implementation details and parameter settings}
To show the effectiveness of the proposed proposal selection algorithm, we implement the proposed CFAPS tracker (Correlation Filter tracking with Adaptive Proposal Selection) based on the original KCFDPT, where the hand-crafted features are used. That is, we concatenate color naming, image intensity and HOG features directly in CFAPS. In addition, we incorporate the adaptive proposal selection algorithm into the HCF tracker~\cite{ma2015hierarchical}. The tracker is named as DeepCFAPS. In the DeepCFAPS tracker, we utilize the outputs of the conv3-4, conv4-4 and conv5-4 convolutional layers from the VGG-Net-19~\cite{simonyan2014very} as features. The values of $\mu_{d}$ of each layer in Eq. (11) are respectively set to 0.25, 0.50 and 1.0, which are similar to~\cite{ma2015hierarchical}.

The regularization parameter $\lambda$ in Eq. (1) and Eq. (2) is set to $10^{-4}$. For the proposal generator, the parameter $s_{d}$ in the detection window size is set to 1.40. The scale penalty parameter $\kappa$ in Eq. (4) is set to 1.40. The damping factor $\beta$ in Eq. (10) is set to 0.70. The above parameters are totally the same as those in KCFDPT. The confidence factor $\eta$ in Eq. (7) and the rate $\eta'$ in Eq. (8) are set to 0.01 and 0.60, respectively. The parameter $\alpha_{D}$ in Eq. (9) is set to 0.15.

% \newcommand{\tabincell}[2]{\begin{tabular}{@{}#1@{}}#2\end{tabular}}
% \begin{table}[t]
% \begin{center}
% \caption{Influence of different proportions of proposals and different instances chosen in the proposal selection algorithm on the OTB2015 dataset} \label{tab:cap}
% \begin{tabular}{|c|c|c|c|}
%   \hline
%   % after \\: \hline or \cline{col1-col2} \cline{col3-col4} ...
%  Name & KCFDPT & CFAPS & \tabincell{c}{Deep\\CFAPS}
%   \\
%   \hline
%   DP(\%) & 74.7 & 77.2 & 83.8 \\
%   AUC(\%) & 57.6 & 58.9 & 59.6 \\
%   Speed(fps)& 40.1& 28.2 & 4.8 \\
%   \hline
% \end{tabular}
% \end{center}
% \end{table}

\newcommand{\tabincell}[2]{\begin{tabular}{@{}#1@{}}#2\end{tabular}}
\begin{table}[t]
\small
\begin{center}
\caption{Analysis of selecting different percentages of proposals on the tracking performance of the OTB2015 dataset. Note that $I_{1}$ or $I_{i-1}$ denotes that only one instance $I_{1}$ or $I_{i-1}$ is used for similarity measurement during tracking.}  \label{tab:cap}
\begin{tabular}{|c|c|c|c|c|c|c|c|}
  \hline
 Percentage(\%) & 30 & 50  & \tabincell{c}{50\\~($I_{1}$)} & \tabincell{c}{50\\~($I_{i-1}$)} & 70 &100
  \\
  \hline
  DP(\%) & 74.3 & \textbf{77.2} & 73.8 & 74.4  & 76.2 & 74.7 \\
  AUC(\%) & 54.1 &\textbf{56.4} & 54.0 & 54.2 & 55.4 & 54.8 \\
  \hline
\end{tabular}
\vspace{-6mm}
\end{center}
\end{table}

%  SV(\%) & 70.6 & 73.1 & 70.9 & 70.9 & 70.9 & 70.3 \\

% \begin{figure}[t]
%  \begin{center}
%  \subfigure[] {\includegraphics[width=0.48\linewidth,height=0.3\linewidth]{./OTB2013PREFastcrop.pdf}}
%  \hspace{0.05in}
%  \subfigure[] {\includegraphics[width=0.48\linewidth,height=0.3\linewidth]{./OTB2015PREFastcrop.pdf}}\\
%  \subfigure[] {\includegraphics[width=0.48\linewidth,height=0.3\linewidth]{./OTB2013PREFastScalecrop.pdf}}
%   \hspace{0.05in}
%  \subfigure[] {\includegraphics[width=0.48\linewidth,height=0.3\linewidth]{./OTB2015PREFastScalecrop.pdf}}\\
%  \subfigure[] {\includegraphics[width=0.48\linewidth,height=0.3\linewidth]{./OTB2013AUCFastcrop.pdf}}
%   \hspace{0.05in}
%  \subfigure[] {\includegraphics[width=0.48\linewidth,height=0.3\linewidth]{./OTB2015AUCFastcrop.pdf}}
% \end{center}
% \vspace{-8mm}
% \caption{Precision and success plots  of CFAPS and the six other state-of-the-art trackers on the OTB2013 dataset (a, e) and OTB2015 dataset (b, f). And precision plots of scale variation on OTB2013 and OTB2015 are shown in (c) and (d), respectively.}
% \label{fig:res}
% % \vspace{-4mm}
% \end{figure}

\begin{figure}[t]
 \begin{center}
 \subfigure[] {\includegraphics[width=0.48\linewidth,height=0.3\linewidth]{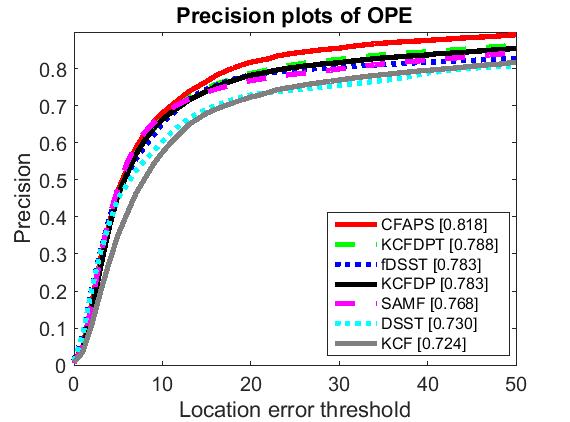}}
 \hspace{0.05in}
 \subfigure[] {\includegraphics[width=0.48\linewidth,height=0.3\linewidth]{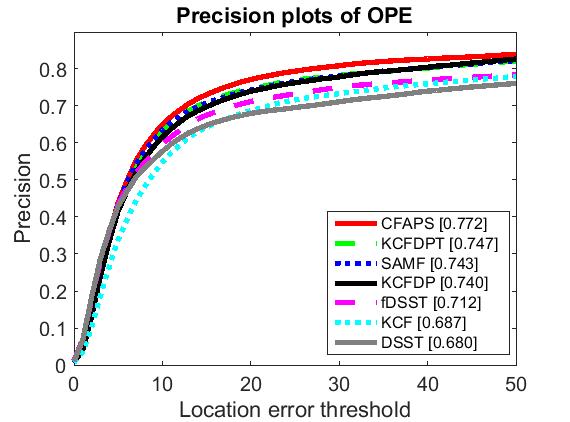}}\\
 \subfigure[] {\includegraphics[width=0.48\linewidth,height=0.3\linewidth]{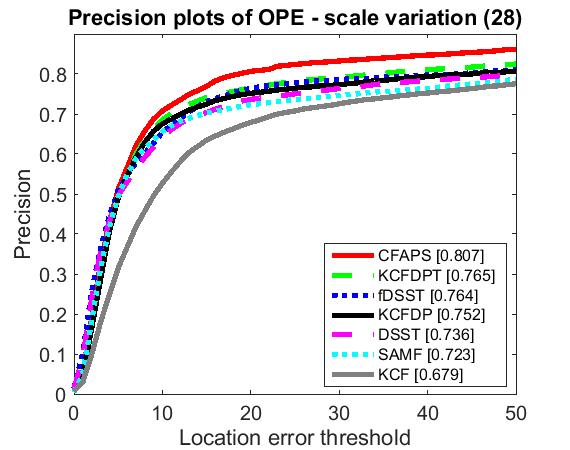}}
  \hspace{0.05in}
 \subfigure[] {\includegraphics[width=0.48\linewidth,height=0.3\linewidth]{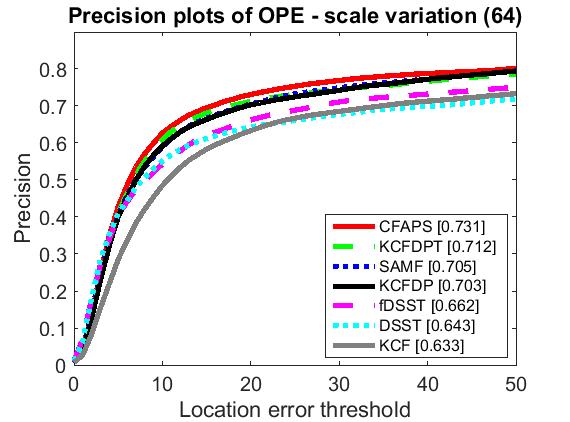}}\\
 \subfigure[] {\includegraphics[width=0.48\linewidth,height=0.3\linewidth]{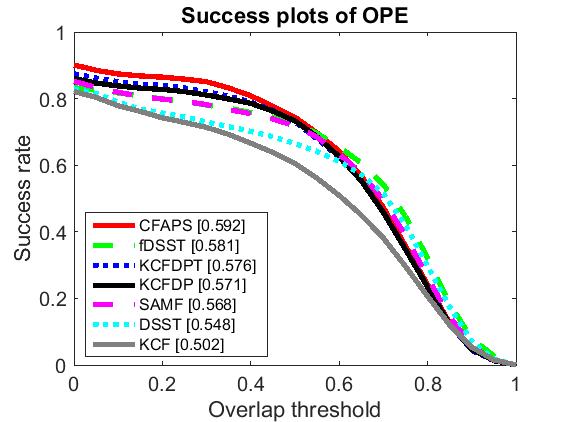}}
  \hspace{0.05in}
 \subfigure[] {\includegraphics[width=0.48\linewidth,height=0.3\linewidth]{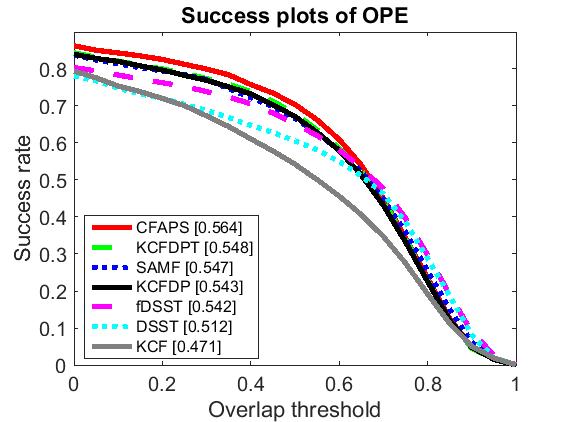}}
\end{center}
\vspace{-9mm}
\caption{Precision and success plots  of CFAPS and the six other state-of-the-art trackers on the OTB2013 dataset (a, e) and OTB2015 dataset (b, f). And precision plots of scale variation on OTB2013 and OTB2015 are shown in (c) and (d), respectively.}
\label{fig:res}
% \vspace{-8mm}
\end{figure}

% The proposed tracker based on hand-crafted features is named "CFAPS"(Kernlized Correlation Filter with Detection Proposals refined by HSV). And the tracker with deep features is named "DeepCFAPS". We analyze the proposed trackers on the OTB2015 Dataset to validate the effectiveness of the Refining proposals and the integration of coarse-to-fine features.
For evaluation metrics, DP (the distance precision at 20 pixels threshold in precision from one pass evaluation (OPE)), and AUC (the area under curve in success plot for OPE) are used in this paper. The OTB 2013 and OTB 2015 datasets are annotated with 11 attributes, including illumination, scale variation, occlussion, deformation, etc. Specifically, We use DP of scale variation to evaluate the tracking performance under the scale variation attribute.

In Table 1, we analyze the impact of selecting different percentages of proposals on the performance of CFAPS on the OTB 2015 dataset. Selecting $100\%$ of proposals corresponds to the KCFDPT tracker. We can see that CFAPS with half of the proposals can achieve the best performance. The results obtained by selecting about $70\%$ and $100\%$ of proposals are slightly inferior to those obtained by selecting about $50\%$ of proposals, which is mainly caused by the distractors contained in the redundant proposals. 
Moreover, we also show the results when only using the initial instance or the previous instance for the proposal selection. The proposed CFAPS tracker can achieve the best performance, which demonstrates the importance of adaptive selection strategy. In the following sections, we will select about $50\%$ of proposals.

\subsection{Comparison with the state-of-the-art algorithms}
We respectively evaluate the performance of CFAPS and DeepCFAPS compared with state-of-the-art algorithms.

\noindent\textbf{Evaluation of CFAPS}. To show that the proposed  adaptive proposal selection algorithm is effective for selecting high-quality proposals to handle scale variations, in this subsection, we compare the CFAPS tracker with several state-of-the-art trackers which are mainly designed for scale variations: SAMF ~\cite{li2014scale}, DSST ~\cite{danelljan2014accurate}, fDSST ~\cite{danelljan2017discriminative}, KCFDPT ~\cite{huang2017applying}, KCFDP ~\cite{huang122015enable} and KCF ~\cite{henriques2015high}.
All experiments are conducted on the Intel I7 3.6GHz CPU. The comparison results are given in Fig.~2.

 CFAPS achieves the top performance among these trackers. Compared with KCF, our tracker outperforms it by a large margin on the DP metric (i.e., 9.4\%/9.2\% on OTB2013 and OTB2015, respectively). On the OTB2013 dataset, CFAPS improves KCFDPT by 2.0\/\%, 2.2\/\% and 1.1\/\% on the DP, DP of scale variation and AUC metrics. On the OTB2015 dataset, CFAPS outperforms KCFDPT by 2.5\/\%, 1.9\/\% and 1.6\/\% on the DP, DP of scale variation and AUC metrics, respectively. Furthermore, KCFDPT runs at 28.6 frames per second (fps), but CFAPS can run at 40.1 fps on average on the OTB2015 dataset, which shows the proposed adaptive proposal selection algorithm can improve the efficiency of the KCFDPT tracker. In general, CFAPS achieves better performance than KCFDPT in terms of tracking accuracy and speed. 
% As far as we know, it is the first online tracker based on the proposal generator.

% % Owing to the factorized convolution operator and sample space model, ECO\_{}HC ranks top on the OTB2013. The scale scheme that ECO adopts is the algorithm of fDSST.
% The results show that the scale algorithm of our tracker performs better than fDSST on OTB2013 and OTB2015. And the algorithm of proposal refining can be easily transplanted to any other correlation trackers and promote the performance efficiently.

%\subsection{}

\noindent \textbf{Evaluation of DeepCFAPS}. We compare the proposed DeepCFAPS with DeepKCFDPT (integrating KCFDPT with HCF~\cite{ma2015hierarchical}) and several state-of-the-art trackers: SAMF ~\cite{li2014scale}, DSST ~\cite{danelljan2014accurate}, fDSST ~\cite{danelljan2017discriminative}, KCFDPT ~\cite{huang2017applying}, KCF ~\cite{henriques2015high}, DeepSRDCF ~\cite{danelljan2015convolutional}, HCF ~\cite{ma2015hierarchical} and LCT ~\cite{ma2015long}. All the experiments are conducted on an NVIDIA GTX TITAN GPU. The comparison results are given in Fig.~3. Note that the HCF tracker in our experiments deploys an extra scale scheme.

As shown in Fig.~3, DeepKCFDPT does not outperform HCF with the scale scheme on the DP and AUC metrics. However, on the OTB2013 dataset, our tracker respectively outperforms the second best tracker (i.e., HCF and LCT) by 1.5\/\% and 1.3\/\% on the DP and AUC metrics. Especially, our tracker outperforms HCF on the DP of scale variation by a large margin of 5.9\/\%. On the OTB2015 dataset, our tracker is not as good as DeepSRDCF on the DP and AUC metrics, but it still outperforms the HCF tracker. Especially, our tracker outperforms DeepSRDCF and HCF by 1.9\/\% and 3.5\/\% on the DP of scale variation metric.

Fig. 4 shows some tracking results obtained by three trackers: KCFDPT, CFAPS and DeepCFAPS. When scale variations are caused by fast motion and motion blur (e.g., Soccer), the KCFDPT tracker chooses those distractors and focuses on the local part of the target. Moreover, when scale variations caused by large deformation occur (e.g., Gym), the bounding boxes of KCFDPT cannot adapt to the appearance of target. Especially, the tracker KCFDPT drifts when the scale variation occurs in Human9. In contrast, the proposed CFAPS and DeepCFAPS trackers can track these targets well.
% \subfigure[] {\includegraphics[width=0.48\linewidth,height=0.3\linewidth]{./OTB2013PREDeepcrop.pdf}}

\begin{figure}[t]
 \begin{center}
 \subfigure[] {\includegraphics[width=0.48\linewidth,height=0.3\linewidth]{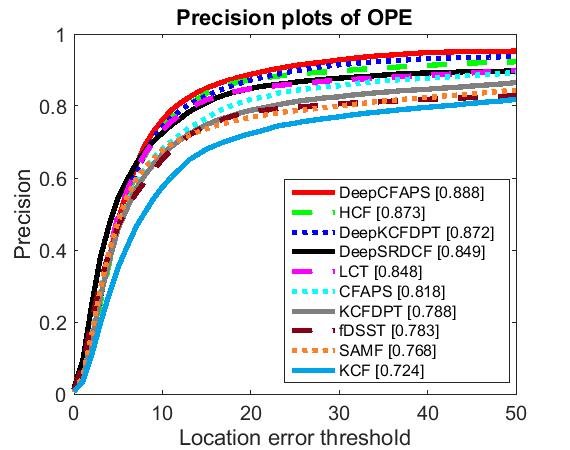}}
 \hspace{0.05in}
 \subfigure[] {\includegraphics[width=0.48\linewidth,height=0.3\linewidth]{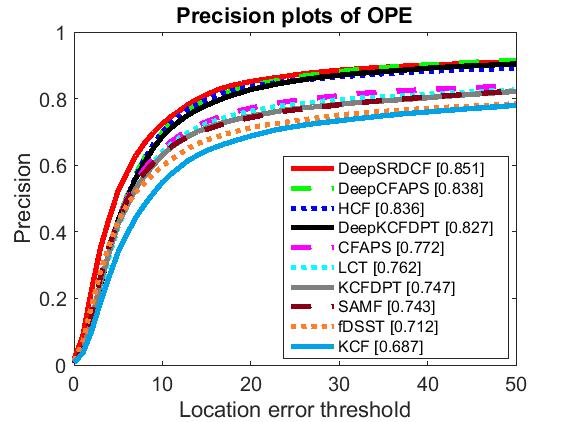}}\\
 \subfigure[] {\includegraphics[width=0.48\linewidth,height=0.3\linewidth]{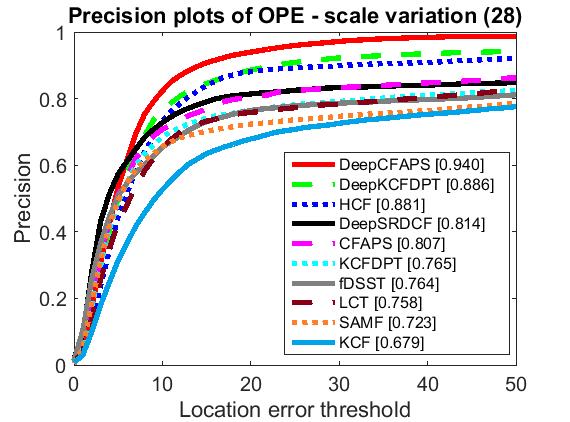}}
  \hspace{0.05in}
 \subfigure[] {\includegraphics[width=0.48\linewidth,height=0.3\linewidth]{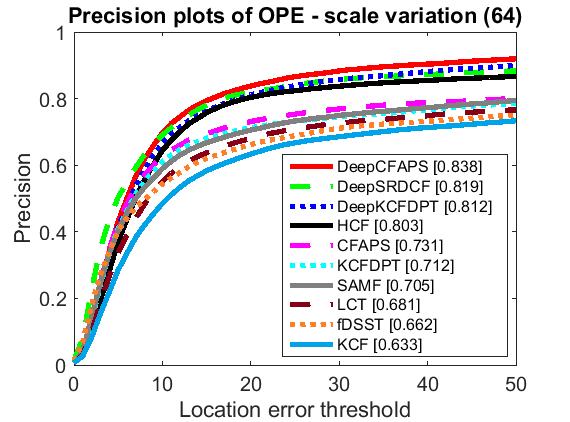}}\\
  \subfigure[] {\includegraphics[width=0.48\linewidth,height=0.3\linewidth]{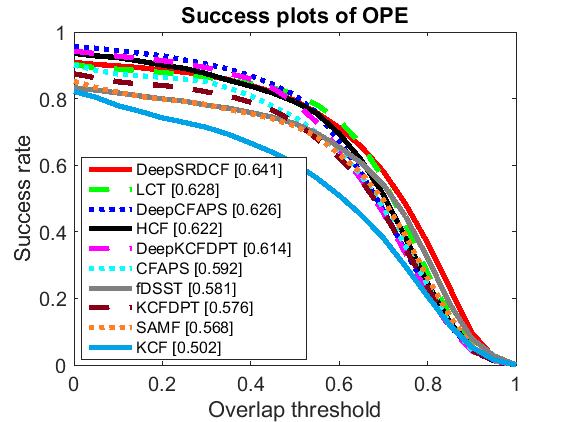}}
 \hspace{0.05in}
 \subfigure[] {\includegraphics[width=0.48\linewidth,height=0.3\linewidth]{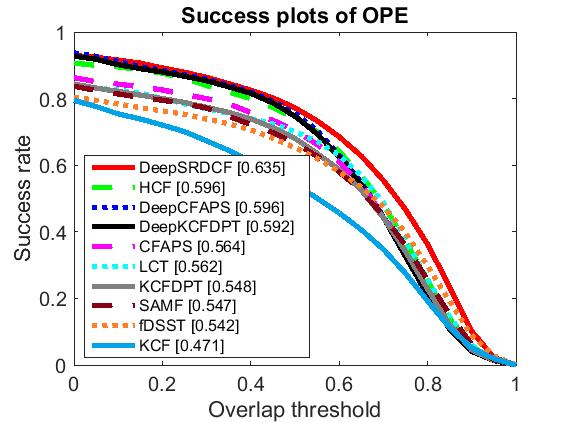}}
\end{center}
\vspace{-8mm}
\caption{Precision and success plots of DeepCFAPS and state-of-the-art trackers on the OTB2013 dataset (a, e) and OTB2015 dataset (b, f).  Precision plots of scale variation on OTB2013 and OTB2015 are shown in (c, d), respectively.}
\label{fig:res}
\vspace{-6mm}
\end{figure}

\begin{figure}[t]
    \centering
    \includegraphics[width=0.9\linewidth,height=0.45\linewidth]{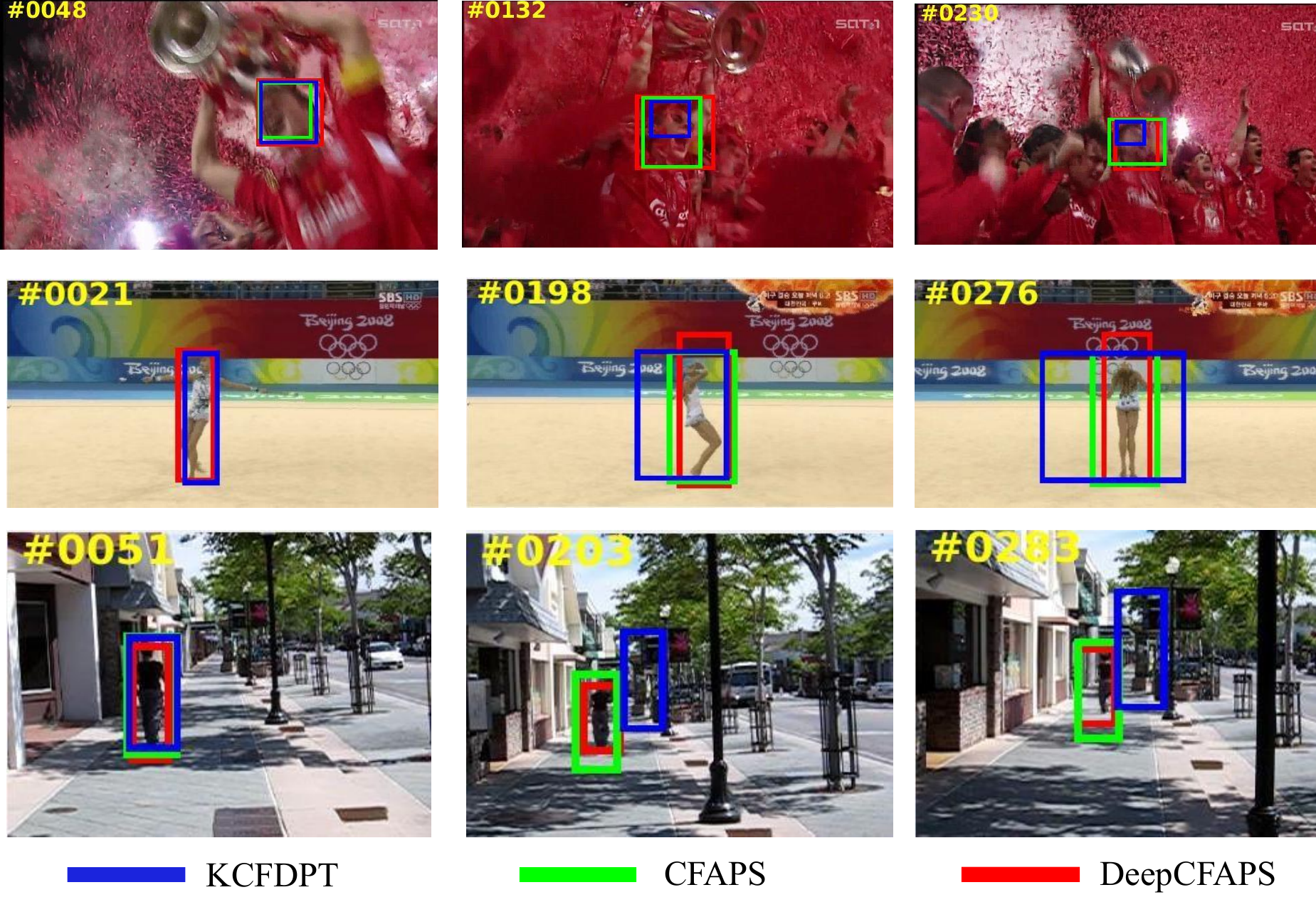}
    \caption{Qualitative evaluation of the trackers KCFDPT, CFAPS and DeepCFAPS on three representative sequences.}
    \label{fig:my_label}
\vspace{-8mm}
\end{figure}

\section{Conclusion}

In this paper, we propose an adaptive proposal selection algorithm for object tracking to effectively handle scale variations. We integrate the proposal selection algorithm with both hand-crafted and deep features to verify the generalization and effectiveness of the algorithm. Extensive experiments on two challenging datasets demonstrate the superiority of the proposed trackers against several state-of-the-art trackers, especially in terms of the scale variations.
\newline

\noindent{\textbf{Acknowledgments.} This work is supported by the National Natural Science Foundation
of China (Grant No. U1605252, 61872307 and 61571379) and the
National Key Research and Development Program of China (Grant
No. 2017YFB1302400).}

%\bibliographystyle{IEEEbib}
%\bibliography{icme2019template}

\end{document}